\documentclass[conference]{IEEEtran}
\IEEEoverridecommandlockouts

\usepackage{hyperref}
\usepackage{url}
\usepackage{booktabs}       
\usepackage{amsfonts}       
\usepackage{nicefrac}       
\usepackage{microtype}      
\usepackage{xcolor}         
\usepackage{multirow}
\usepackage{natbib}
\usepackage{wrapfig}
\usepackage{tablefootnote}
\usepackage{caption}
\usepackage{subcaption}
\usepackage{graphicx}
\usepackage{textcomp}
\usepackage{amsmath}
\usepackage{colortbl}
\usepackage[h]{esvect}
\usepackage{pifont}
\usepackage{tablefootnote}
\usepackage{enumitem}
\usepackage{pifont}
\usepackage{fancyhdr}

\DeclareMathOperator*{\argmin}{arg\,min}

\def\BibTeX{{\rm B\kern-.05em{\sc i\kern-.025em b}\kern-.08em
    T\kern-.1667em\lower.7ex\hbox{E}\kern-.125emX}}

\definecolor{ndssblue}{RGB}{0,61,165} 

\fancypagestyle{firstpage}{
    \fancyhf{} 
    \fancyhead[C]{%
        \textcolor{ndssblue}{\Large \textbf{This paper has been accepted to IEEE CIC 2025}}%
    }
}

\begin{document}

\title{Maximizing Information in Domain-Invariant Representation Improves Transfer Learning}

\author{
\IEEEauthorblockN{Adrian Shuai Li\IEEEauthorrefmark{1},
Elisa Bertino\IEEEauthorrefmark{1},
Xuan-Hong Dang\IEEEauthorrefmark{2},
Ankush Singla\IEEEauthorrefmark{1},
Yuhai Tu\IEEEauthorrefmark{3},
Mark N. Wegman\IEEEauthorrefmark{2}}

\IEEEauthorblockA{\IEEEauthorrefmark{1}Department of Computer Science, Purdue University, West Lafayette, IN, USA \\
Email: \{li3944, bertino, asingla\}@purdue.edu}

\IEEEauthorblockA{\IEEEauthorrefmark{2}IBM T. J. Watson Research Center, Yorktown Heights, NY, USA \\
Email: \{xuan-hong.dang, wegman\}@us.ibm.com}

\IEEEauthorblockA{\IEEEauthorrefmark{3}Center for Computational Biology and Center for Computational Neuroscience, \\
Flatiron Institute, New York, NY, USA\\
Email: ytu@flatironinstitute.org}
}

\maketitle
\thispagestyle{firstpage}  

\begin{abstract}
 
We propose MaxDIRep, a domain adaptation method that improves the decomposition of data representations into domain‑independent and domain‑dependent components. Existing methods, such as Domain‑Separation Networks (DSN), use a weak orthogonality constraint between these components, which can lead to label‑relevant features being partially encoded in the domain‑dependent representation (DDRep) rather than the domain‑independent representation (DIRep). As a result, information crucial for target‑domain classification may be missing from the DIRep. MaxDIRep addresses this issue by applying a Kullback‑Leibler (KL) divergence constraint to minimize the information content of the DDRep, thereby encouraging the DIRep to retain features that are both domain‑invariant and predictive of target labels. Through geometric analysis and an ablation study on synthetic datasets, we show why DSN’s weaker constraint can lead to suboptimal adaptation. Experiments on standard image benchmarks and a network intrusion detection task demonstrate that MaxDIRep achieves strong performance, works with pretrained models, and generalizes to non‑image classification tasks.
\end{abstract}


 \section{Introduction}

Domain adaptation (DA) tackles the challenge of training classifiers for a target domain with limited or no labels by leveraging labeled data from a related source domain. This setting is valuable because obtaining labeled data can be costly and time‑consuming, particularly in real‑world applications such as image recognition and network intrusion detection~\cite{singla2020preparing}.

Neural networks often exploit contextual cues, such as background, during training~\cite{mitAvoidingShortcut}. For example, wolves are frequently depicted in wild environments, whereas huskies are not. In a DA scenario where the source domain contains wolves and huskies in their natural habitats and the target domain shows them in veterinary clinics, these background cues become irrelevant. Such domain‑specific cues, or “spurious correlations,” can hinder cross‑domain generalization because the features learned during training may not transfer well to the target domain. An effective DA method must therefore produce representations that are both domain‑invariant and sufficiently informative for target label prediction.

Our general intuition, largely consistent with previous work~\citep{bousmalis2016domain, stojanov2021domain}, is that effective DA requires two conditions:
\begin{enumerate}
\item A representation of the input is formed that is independent of the domain; we call this a domain-independent representation (DIRep).
\item The DIRep contains all the information relevant for classification in the target domain.
\end{enumerate}

A common approach to achieving the first condition is to use adversarial techniques such as generative adversarial networks (GANs)~\citep{ganin2016domain,singla2020preparing, tzeng2017adversarial}.  These techniques ensure that the DIRep discloses no information about the data's original domain. However, GANs alone do not guarantee that the learned DIRep contains information relevant for predicting the labels in the target domain~\citep{stojanov2021domain}.

To satisfy the second condition, one strategy is to encode all data information (from both domains) into the representation using an autoencoder. This cannot be done with the DIRep alone, as it should not contain domain-dependent information. To address this, a domain-dependent representation (DDRep) is introduced, as in Domain-Separation Networks (DSN)~\citep{bousmalis2016domain}. The data is then represented by both the DIRep and the DDRep, which are used together to reconstruct the data in the autoencoder. We adopt this approach, but with a different method of decomposing the DDRep and DIRep than DSN.

A key challenge in DA is determining the optimal partitioning of information between the DIRep and the DDRep. Unlike DSN, which only enforces orthogonality between these representations, our approach, MaxDIRep, explicitly minimizes the information content of the DDRep by constraining it with a Kullback-Leibler (KL) divergence to a standard normal distribution. This constraint ensures that the DIRep captures as much relevant information as possible (consistent with our first condition, achieved via adversarial training). By minimizing the DDRep information content, we ensure that only domain identity, which is irrelevant for classification, is encoded within it. This contrasts with DSN, where useful target domain information can reside in the DDRep, hindering classification performance. 

The contributions of this study are summarized as follows.

\begin{itemize}
\item We introduce MaxDIRep, a method using a KL divergence constraint on DDRep to minimize its information content, ensuring relevant classification information resides in DIRep. We provide theoretical intuition through a geometric analysis and empirical evidence via an ablation study.
\item Using synthetic benchmarks with domain‑specific cues, we show that MaxDIRep achieves the lowest error rate for an ideal joint hypothesis, confirming that its DIRep better captures target‑label‑relevant information. Results are consistent across two synthetic benchmarks.

\item MaxDIRep demonstrates comparable or superior performance to other recent DA methods on standard DA benchmarks (Office-31 and Office-Home).
\item We demonstrate MaxDIRep’s applicability beyond image tasks by improving network intrusion detection performance over existing DA‑based baselines.
\end{itemize}

Our code is available \href{https://github.com/gloryer/MaxDIRep/tree/main}{here} to support future research.

\section{Related Work}\label{related}

Transfer learning is an active research area covered by several survey papers~\citep{liu2022deep,zhang2022transfer,zhang2021survey,zhuang2020,liu2019transferable,wang2018deep}. We briefly describe previous methods closely related to our work.

  The domain adversarial neural network (DANN)~\citep{ganin2016domain} uses a generator, a label predictor, and a domain classifier.  The generator is trained at the same time as the label predictor, which takes the generator's output as its input to create a DIRep that contains features for labels. It is also trained in an adversarial fashion to ensure domain-dependent information doesn't get into its output by reversing the loss function of the domain classifier. The adversarial discriminative domain adaptation (ADDA)~\citep{tzeng2017adversarial} uses similar network components with a learning process that involves multiple stages in training the three components of the model. Singla et al.~\cite{singla2020preparing} have proposed a hybrid version of the DANN and ADDA where the generator is trained with the standard GAN loss function~\citep{goodfellow2020generative}; we refer to this as the Singla method~\citep{singla2020preparing}.  None of these methods (DANN, ADDA, and Singla method) includes an autoencoder and thus does not have a DDRep.

The Domain-Specific Adversarial Network (DSAN)~\citep{stojanov2021domain} uses domain-specific information as input to the encoding function, in addition to the data, to infer the DIRep. In contrast, our approach learns the DIRep without incorporating domain-specific information as input. The closest approach to ours is Domain-Separation Networks (DSN)~\citep{bousmalis2016domain}. The key distinction between DSN and our method lies in the constraints used for decomposing the data representation into DIRep and DDRep. DSN uses a ``soft subspace orthogonality constraint between the private and shared representation of each domain'' to ensure distinct DIRep and DDRep components, which have the same shape. 
Cai et al.~\cite{cai2019learning} train their equivalent of the DDRep using an adversarial network to ensure that the DDRep does not contain any information that can be useful for classification. Our approach uses a stronger constraint to minimize the DDRep's information content. Details are provided in Section~\ref{compare}. Since we don't test linear orthogonality, we don't require that the DDRep and DIRep have the same shape.

Other work explores leveraging multiple target~\citep{peng2019domain} or source domains~\citep{pei2018multi,park2021information}. Some authors have evaluated cross-domain representation disentanglement on image-to-image translation and retrieval, such as the Interaction Information Auto-Encoder (IIAE)~\citep{hwang2020variational}. The Variational Disentanglement Network (VDN)~\citep{wang2022variational} aims to generalize from a source domain without access to a target. These methods either address different problem settings (instead of one source domain and one target domain)~\citep{peng2019domain, pei2018multi,park2021information,wang2022variational} or use non-adversarial training for extracting domain-invariant features~\citep{hwang2020variational}. As these works are less relevant to our approach, we do not discuss them further.

\section{The MaxDIRep Model}
\label{Detailed Design}

\begin{figure}[h]
\centering
\includegraphics[width=0.85\linewidth]{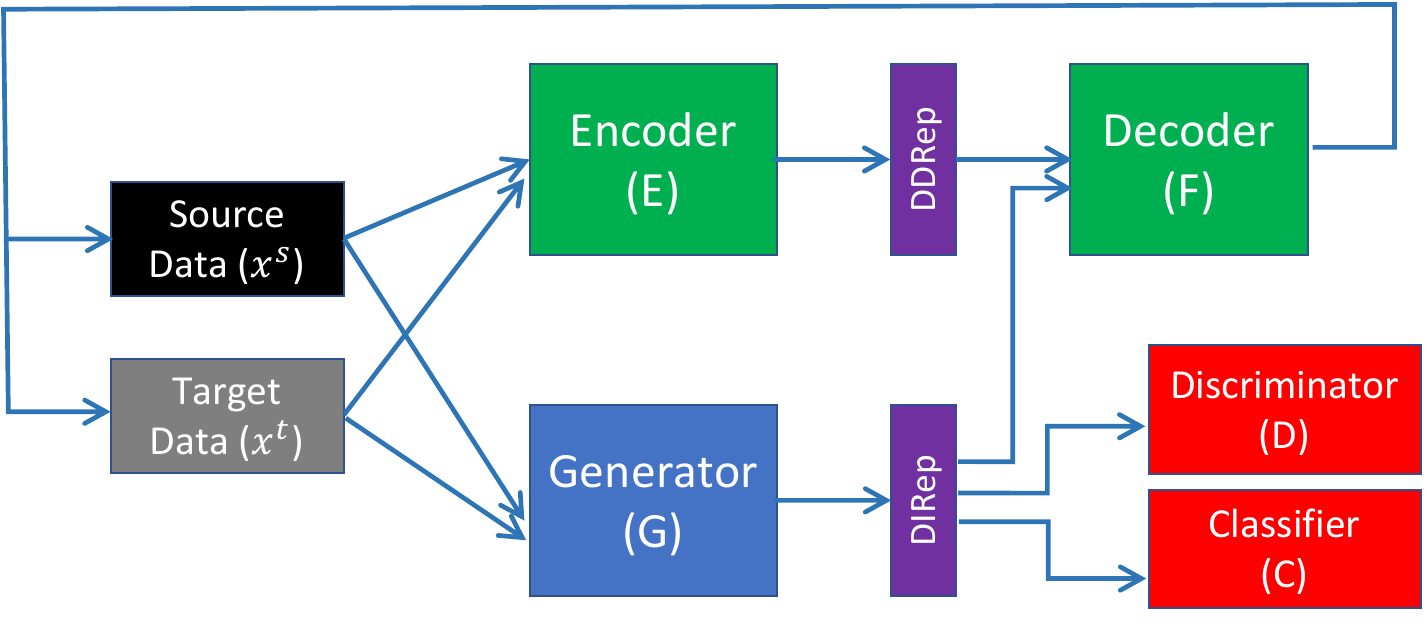}
\caption{Architecture of MaxDIRep.}
\label{illu}
\vspace{-2mm}
\end{figure}

This section details our method, MaxDIRep (summarized in Figure~\ref{illu}). To achieve effective adaptation, our goal is to constrain DIRep extraction to retain the maximal information about the target labels. MaxDIRep achieves this by minimizing the information content of the DDRep during data generation from both DDRep and DIRep. We measure the KL divergence between the DDRep and a standard normal distribution (a baseline distribution with minimal information). Including this KL divergence in the overall loss function constrains the DDRep's information content. By minimizing domain-specific information in the DDRep, we force the DIRep to retain maximal information relevant to target labels.  DIRep is also subject to a GAN-like discriminator, which ensures domain-invariant classification information.

\subsection{Loss functions and model training}

\subsubsection{Networks} There are five neural networks (by neural network, we mean the network architecture and all its parameters) in the algorithm: \ding{182} $G$ is the generator; \ding{183} $D$ is the discriminator;  \ding{184} $C$ is the classifier;  \ding{185} $E$ is the encoder;  \ding{186} $F$ is the decoder.

\subsubsection{Inputs and outputs} The data is represented as $(x, l, d)$, where $x$ is the input, $l$ is the label of sample $x$ (if available), and $d$ indicates the domain identity (i.e., a single bit, 0 for the source domain and 1 for the target domain). We use $x^s$ and $x^t$ to denote source and target data samples, respectively, when necessary. In zero-shot or few-shot DA settings, $l$ is available for all source data samples, but no or only a few labels are available for the target samples. The input $x$ is provided to both the encoder ($E$) and the generator ($G$). The DDRep and DIRep correspond to the intermediate outputs of $E$ and $G$, respectively:
 \begin{equation*}
 \mathit{DDRep=E(x)},\;\;\; \mathit{DIRep=G(x)}
 \end{equation*}
which then serve as the inputs for the downstream networks: decoder ($F$), discriminator ($D$), and classifier ($C$). In particular, DIRep serves as input for $D$ and $C$, and both DIRep and DDRep serve as the inputs for $F$. The outputs of these three downstream networks are $\hat{x}$ from the decoder $F$,  $\hat{d}$ from the discriminator $D$, and $\hat{l}$ from the classifier $C$:
\begin{equation*}
    \mathit{\hat{x} =F(E(x),G(x))},\;\;
    \mathit{\hat{d} =D(G(x))},\;\; \mathit{\hat{l}= C(G(x))}
\end{equation*} 
where we explicitly list the dependence of the outputs on the corresponding networks.

\subsubsection{Loss functions} 
First, we introduce all the loss functions designed in MaxDIRep to achieve effective adaptation. We then demonstrate how these loss functions are integrated into a joint training framework.

In unsupervised DA, the classification loss applies only to the source domain, and it is defined as follows:
\begin{equation}
\mathcal{L}_c= 
 -  \sum_{i=1}^{N_s} {l}_i^s \cdot log \hat{l}_i^s
 \end{equation}%
where $N_s$ represents the number of samples from the source domain, $l_i^s$ is the one-hot encoding of the label for the source input $x_i^s$, and $\hat{l}_i^s$ is the softmax output of $C(G(x_i^s))$. 

The discriminator loss trains the discriminator to predict whether the DIRep is generated from the source or the target domain. $N_t$ represents the number of samples from target domain and $\hat{d}_i$ is the output of $D(G(x_i))$.
\begin{equation}
\mathcal{L}_d= 
 -  \sum_{i=1}^{N_s + N_t} \left\{ {d}_ilog \hat{d}_i + (1-{d}_i)log (1-\hat{{d}}_i) \right\} 
\end{equation}%

The generator loss is the GAN loss with inverted domain truth labels:
\begin{equation}
\mathcal{L}_g= 
 -  \sum_{i=1}^{N_s + N_t} \left\{ (1-{d}_i)log \hat{d}_i + d_ilog (1-\hat{{d}}_i) \right\} 
\end{equation}%

For the reconstruction loss, we use the standard mean squared error loss calculated from both domains:
\begin{align}
    \mathcal{L}_r = &\ \sum_{i}^{N_s}||x_i^s - \hat{x}_i^s||_2^2 + \sum_{i}^{N_t}||x_i^t - \hat{x}_i^t||_2^2 
\end{align}%
where $\hat{x}_i^s = F(G(x_i^s), E(x_i^s))$ and $\hat{x}_i^t = F(G(x_i^t), E(x_i^t))$

The losses $\mathcal{L}_c$, $\mathcal{L}_d$, $\mathcal{L}_g$, and $\mathcal{L}_r$ are analogous to those used in other GAN-based DA algorithms such as DSN. Adversarial training combined with the source‑only classification loss ensures that the DIRep does not disclose domain information. However, this does not guarantee that DIRep contains all features relevant to target‑domain classification. The key distinguishing feature of our proposed method is the KL divergence loss $\mathcal{L}_{kl}$ applied to the DDRep. The inclusion of $\mathcal{L}_{kl}$ aims to induce a DDRep with minimal information content, thereby necessitating a more prominent role of the DIRep during data reconstruction. This, in turn, forces the DIRep to include sufficient information for effective target domain classification.

To this end, we measure the KL divergence between the distribution of DDRep and a standard normal distribution, which serves as a baseline distribution with minimal information.   We assume that DDRep follows a normal distribution with mean $\mathbb{E}(DDRep)$ and variance $\mathbb{V}(DDRep)$, $\mathcal{L}_{kl}$ is defined as:
\begin{equation}
\begin{split}
    \mathcal{L}_{kl} &= \mathit{D_{KL}(DDRep\parallel\mathcal{N}(0,I) )}\\ 
    &=  -\frac{1}{2}(1 + \mathit{log[\mathbb{V}(DDRep)]}  
    -  \mathit{\mathbb{V}(DDrep)}  - \mathit{\mathbb{E}(DDRep)^2)}
\end{split}
\end{equation}

\subsubsection{The back-prop based learning} The gradient descent-based learning dynamics for updating the five neural networks are described by the following equations:
\begin{equation*}
\begin{aligned}
\Delta G &= -\alpha_G \left(\lambda \frac{\partial \mathcal{L}_g}{\partial G} + \beta \frac{\partial \mathcal{L}_c}{\partial G} + \gamma \frac{\partial \mathcal{L}_r}{\partial G}\right), \\
\Delta C &= -\alpha_C \frac{\partial \mathcal{L}_c}{\partial C}, \quad 
\Delta D = -\alpha_D \frac{\partial \mathcal{L}_d}{\partial D}, \\
\Delta E &= -\alpha_E \left(\frac{\partial \mathcal{L}_{kl}}{\partial E} + \mu \frac{\partial \mathcal{L}_r}{\partial E}\right), \\
\Delta F &= -\alpha_F \frac{\partial \mathcal{L}_r}{\partial F}
\end{aligned}
\end{equation*}
where $\alpha_{C,D,E,F,G}$ are the learning rates for different neural networks. In our experiments, we often set them to the same value, but they can be different in principle. The other hyperparameters, namely $\lambda$, $\beta$, $\gamma$, and $\mu$, are the relative weights of the loss functions. 

\subsection{The explicit DDRep model}
From the results of the  MaxDIRep algorithm, we observed that the DDRep contains a small amount of information, as measured by the KL divergence (consistently small across all experiments; see Table~\ref{appkl} in the Appendix). Motivated by this observation, we introduce a simplified MaxDIRep algorithm without the encoder $E$, where the DDRep is explicitly set to the domain label (bit) $d$, i.e., $\mathit{DDRep=d.}$  We refer to this simplified algorithm as the explicit DDRep algorithm. The rationale is that $d$ represents the simplest possible domain-dependent information. 

Beyond its simplicity, the explicit DDRep algorithm offers high interpretability. A particularly useful feature is that it allows us to directly examine the effect of the DDRep by flipping the domain bit ($d\rightarrow 1-d$).
If the reconstructed image $\mathit{\tilde{x} = F(DIRep, 1-d)}$  resembles an image from the other domain, we can infer that the domain bit effectively captures domain-dependent information (see Section~\ref{Cheating} for details and Figure~\ref{recon_fm} for examples of reconstructed images).

In our experiments, the explicit DDRep algorithm performs comparably to the MaxDIRep model in some simple cases (see Section~\ref{Cheating}). However,  MaxDIRep performs better in more complex scenarios (Sections~\ref{Standard} and~\ref{nid}). Therefore, we use the  MaxDIRep model with $\mathcal{L}_{kl}$ for all cases except the experiments in Section~\ref{Cheating}, where the explicit DDRep algorithm performs equally well and provides direct interpretability.

\subsection{Comparative analysis of MaxDIRep and DANN: insights from DA theory } 

We now provide a theoretical analysis of MaxDIRep based on the DA theory established in Theorem 1 of \cite{ben2010theory}. While deriving the explicit target error bound for MaxDIRep turns out to be formidable, we provide some insights into why MaxDIRep yields better adaptability than DANN, grounded in Theorem 1. These insights will be empirically validated through experiments in Section~\ref{Cheating}.

\textbf{Theorem 1.} (Ben-David et al.~\cite{ben2010theory}). Let $\mathcal{H}$ be the hypothesis space and $\mathcal{E}_s (h)$, $\mathcal{E}_t (h)$ be the error of hypothesis $h \in \mathcal{H}$  on the source domain $X_s$ and the target domain $X_t$, respectively. Then for any classifier $h \in \mathcal{H}$, the error on the target domain is bounded by,
 \begin{equation}\label{t1}
 \mathcal{E}_t (h) \leq \mathcal{E}_s (h) + d_{\mathcal{H}\Delta\mathcal{H}} (X_s, X_t) + \lambda, 
 \end{equation}

 where $d_{\mathcal{H}\Delta\mathcal{H}}$ is the $\mathcal{H}\Delta\mathcal{H}$ distance measuring domain shift and $\lambda$ is the error of an ideal joint hypothesis defined as $h^{*} = \argmin_{h\in \mathcal{H}}\mathcal{E}_s(h) + \mathcal{E}_t(h) $, such that  
 \begin{equation}
 \lambda = \mathcal{E}_s(h^{*}) + \mathcal{E}_t(h^{*}) 
 \end{equation}

  In DANN, training the discriminator on the DIRep bounds the $\mathcal{H}\Delta\mathcal{H}$ distance while training the feature extractor and the classifier on the source labeled data minimizes the error on the source domain ($\mathcal{E}_s (h)$) (see the proof in \cite{ganin2016domain}). The third term, $\lambda$, is assumed to be sufficiently small in their analysis. However, as previous work has shown~\citep{chen2019transferability, liu2019transferable}, the error of the ideal joint hypothesis $h^*$, especially for the target domain $\mathcal{E}_t(h^{*})$, cannot be overlooked in DANN. We present a reasonable explanation for this.  In an unsupervised DA task, where the target data lacks labels, the classifier can use source-specific information that helps with source classification. Consequently, information that could be beneficial for classifying the target data may be omitted from the DIRep, leading to an increased $\mathcal{E}_t(h^{*})$.
 
MaxDIRep addresses this issue by (1) decomposing the full representation into DIRep and DDRep, ensuring their combination contains sufficient information for data reconstruction; (2) aligning the DDRep distribution with a standard normal distribution to minimize its information content; and (3) ensuring that the DIRep 
is domain invariant subject to adversarial training. By doing so, the DIRep can capture more relevant domain-invariant features useful for target classification, as the information in DDRep is minimized. The improved target representation can reduce the generalization error on the target domain, thus reducing $\mathcal{E}_t(h^{*})$. Consequently, the $\lambda$ term is further bounded, yielding a tighter bound for  $\mathcal{E}_t (h)$ than DANN. We will justify this in Section~\ref{Cheating} (See Figure~\ref{error} for the error rate of an ideal joint hypothesis trained using representations learned by DANN, DSN, and MaxDIRep).

\subsection{Comparative analysis of MaxDIRep and DSN: MaxDIRep has a stronger constraint than DSN } \label{compare}
To better understand the differences between DSN and MaxDIRep, we next formalize their constraints mathematically.
Both DSN and MaxDIRep decompose the data representation into DIRep and DDRep.  The main difference\footnote{DSN also uses different neural networks to create the DDRep from their source and target.} is that instead of using $\mathcal{L}_{kl}$ to force the DDRep to contain minimal information as in MaxDIRep, DSN uses a linear orthogonality constraint between the private and shared representations of each domain. Formally, this constraint ($\mathcal{L}_\mathit{{diff}}$) is enforced by minimizing the dot product between the DDRep ($DD^{S/T}$) and DIRep ($DI$) for both source ($S$) and target ($T$) data:
\begin{equation}
\mathit{L_{diff}= \left\Vert DI \cdot DD^{S}\right\Vert^2 + \left\Vert DI\cdot DD^{T}\right\Vert^2}
\end{equation}

However, the orthogonality constraint does not always lead to a unique and optimal decomposition.  For instance, an alternative, yet still (nearly) orthogonal decomposition would minimize the information content in the DIRep,  with most image details contained in the DDRep.  As discussed in Section~\ref{DIRep_max}, this decomposition leads to poor DA performance but is not precluded by the DSN algorithm due to its weaker linear orthogonality constraint.

\begin{figure}[t]
\centering
\includegraphics[trim=150 210 145 220, clip, width=1.1\linewidth]{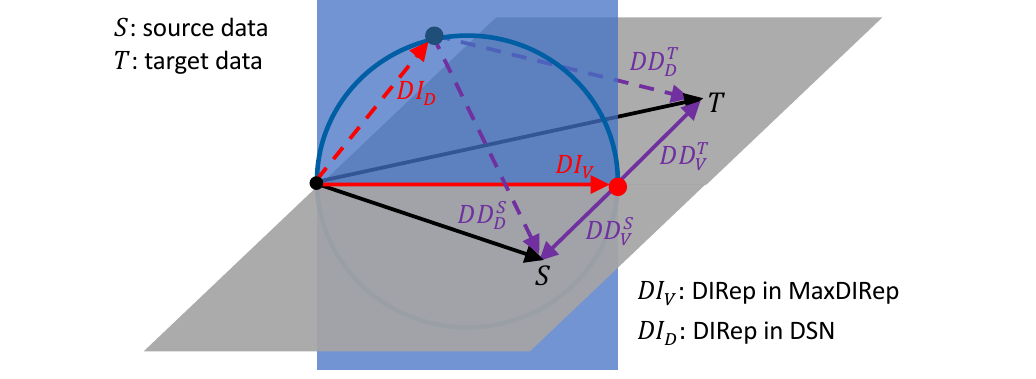}
\caption{Schematic comparison between DSN and MaxDIRep (best viewed in color). See the text for details.}
\label{DA_Geom}
\vspace{-2mm}
\end{figure}

To gain intuition about the difference between DSN and MaxDIRep, we consider a 3-D geometrical analogy of a representation decomposition as shown in Figure~\ref{DA_Geom}.  In this analogy, source ($S$) and target ($T$) data, represented by vectors in 3D space, are decomposed into the sum of DIRep ($DI$) and DDRep ($DD$):
\[S = DI_x + DD^S_x,\ T = DI_x + DD^T_x\]
where the subscript $x$ denotes either the DSN ($D$) or MaxDIRep ($V$) algorithm. In DSN, the linear orthogonality constraint, $DI_D\cdot DD_D^{S,T}=0$, enforces $\mathit{DI_D\perp DD_D^{S,T}}$, which is satisfied by any points on the blue circle in Figure~\ref{DA_Geom}. In MaxDIRep, however, the magnitude of the DDRep, i.e., $||S - DI|| + ||T - DI||$, is minimized, resulting in a unique solution $DI_V$ (the red dot in Figure~\ref{DA_Geom}). \textit{Our solution not only satisfies the orthogonality constraint ($DI_V \perp DD^{S,T}_V$) but also maximizes the magnitude of the DIRep ($||DI_V|| \geq ||DI_D||$) - see Appendix~\ref{geom} for proof.}

This 3D geometric analogy suggests that the orthogonality constraint is weaker than minimizing the magnitude of the DDRep. Depending on the initialization, a system relying solely on the orthogonality constraint can converge to a suboptimal solution (any point on the circle other than the MaxDIRep solution $DI_V$) with inferior DA performance. For instance, as illustrated in Figure~\ref{DA_Geom}, the origin, i.e., $DI_D = 0$, represents a valid solution for DSN that satisfies the orthogonality constraint. This extreme case, with a minimal (zero) DIRep, is unsuitable for adaptation.

We anticipate that the DA performance will degrade as the DSN solution deviates further from the MaxDIRep solution. Indeed, as demonstrated in Section~\ref{DIRep_max} through a series of ``mutual ablation'' experiments in a realistic setting, perturbing the DSN system by applying a KL loss $\mathcal{L}_{\mathrm{kl}}^{DI}$ to its DIRep for a certain time causes DSN to converge to solutions consistent with its orthogonality constraint but exhibiting poorer DA performance. Moreover, increasing the strength of this perturbation further degrades DSN's performance, indicating the existence of many suboptimal solutions consistent with the geometric analogy. However, the opposite is not true, i.e., perturbing the MaxDIRep system by applying a negative $\mathcal{L}_{\mathrm{diff}}$ to make the DIRep and DDRep less orthogonal does not prevent MaxDIRep from converging to the optimal solution with comparable DA performance.

 \section{Experiments}
 \label{Experiments}

We evaluate MaxDIRep across various adaptation settings. In Section~\ref{Cheating} and \ref{synthetic_cifar10}, we use synthetic datasets to explicitly demonstrate MaxDIRep's advantage over DANN and DSN, which can exploit source-specific information, leading to suboptimal DA performance. Specifically, we introduce ``cheating information''—spurious correlations that aid source domain classification but are ineffective in the target domain. This ``cheating information'' can bias the learned DIRep to lack sufficient information for target label prediction, resulting in poor DA performance.

Next, in Section~\ref{DIRep_max}, we conduct a series of mutual ablation experiments between MaxDIRep and DSN to demonstrate that MaxDIRep's superior performance stems from its stronger constraint on minimizing the DDRep's information content compared to DSN's orthogonality constraint.

In Section~\ref{Standard}, we compare the performance of MaxDIRep on standard DA benchmark datasets, including Office-31~\citep{saenko2010adapting} and Office-Home datasets~\citep{venkateswara2017deep}. Although the primary focus of this work is to compare our method with DANN and DSN, we also include comparisons with several recent methods on these datasets to illustrate the practical value of our approach. Overall, our approach achieves comparable or superior results on standard DA benchmarks.

Finally, in Section~\ref{nid}, we demonstrate MaxDIRep's application in training network intrusion detectors, building on the Singla method~\cite{singla2020preparing}, which addressed label scarcity in this domain using DA. Our results show that MaxDIRep consistently improves upon the Singla method and outperforms DSN and DANN, highlighting MaxDIRep's versatility for non-image classification tasks.

\subsection{Synthetic benchmark based on Fashion-MNIST}\label{Cheating}

Using Fashion-MNIST as the source domain, we create a target domain by rotating the images by 180 degrees. The core idea is to introduce ``cheating information'' that allows perfect source domain classification but hinders target domain generalization. To achieve this, we append a one-hot label vector (``cheating bits'') to each flattened source image (reshaped into a $1 \times N$ vector, where $N$ is the number of pixels). We also append bits to the target images, but these are not the true target labels. We consider two approaches for generating these target ``cheating bits'': random label assignment (random cheating) and shifting the true label by one index (shift cheating).  The cheating bits in the target data have the same distribution as those in the source data. This setup ensures that if an algorithm relies on the ``cheating bits,'' it will perform well on the source domain but poorly on the target domain, effectively demonstrating the problem we aim to address.

\subsubsection{Benchmark algorithms and results}

We compare MaxDIRep with three adversarial DA algorithms: the Singla method~\citep{singla2020preparing}, DANN~\citep{ganin2016domain} and DSN~\citep{bousmalis2016domain}. We implemented both MaxDIRep and the explicit DDRep algorithm in the zero-shot setting. The explicit DDRep algorithm and MaxDIRep achieve almost identical performance. We also provide two baselines: a classifier trained on the source domain samples without DA (which gives us the lower bound on target classification accuracy) and a classifier trained on the target domain samples (which gives us the upper bound on target classification accuracy). More details of the topology, learning rate, and hyperparameter setup are provided in Appendix~\ref{appfm}.

We compare the mean accuracy of our approach and the other DA algorithms on the target test set in Table~\ref{fmnist}.  The z-scores, which indicate the statistical significance of the performance difference between our method and others, are shown in Table~\ref{appz test FM}. 
In the no-cheating scenario, MaxDIRep outperforms the Singla method, DANN, and achieves comparable performance to DSN. The Singla method and DANN  experience a $5\%$ accuracy drop with shift cheating and a $10\%$ drop with random cheating. In contrast, our method exhibits only a $0.1\%$ and  $5\%$ accuracy drop, respectively. While DSN performs better than the Singla method and DANN in the presence of cheating bits, our approach still significantly outperforms DSN in both the shift and random cheating scenarios. 

\begin{table}[h]

\centering
\caption{Mean classification accuracy (\%) of different adversarial learning-based DA approaches on the synthetic Fashion-MNIST benchmark.}
\resizebox{\linewidth}{!}{%
\begin{tabular}{llll}
\hline
Model &
  \begin{tabular}[c]{@{}r@{}}No\\ cheating\end{tabular} &
  \begin{tabular}[c]{@{}r@{}}Shift\\ cheating\end{tabular} &
  \begin{tabular}[c]{@{}r@{}}Random\\ cheating\end{tabular} \\ \hline
Source-only & \multicolumn{1}{r}{20.0}    & \multicolumn{1}{r}{11.7}  & \multicolumn{1}{r}{13.8}   \\
Singla method~\citep{singla2020preparing}  &    \multicolumn{1}{r}{64.7}                      &    \multicolumn{1}{r}{58.2}                      &      \multicolumn{1}{r}{54.8}                        \\
DANN~\citep{ganin2016domain}        & \multicolumn{1}{r}{63.7} & \multicolumn{1}{r}{58.0} & \multicolumn{1}{r}{53.6} \\
DSN~\citep{bousmalis2016domain}        &     \multicolumn{1}{r}{66.8 }                    &       \multicolumn{1}{r}{63.6 }                  &       \multicolumn{1}{r}{57.1}                   \\
MaxDIRep/Explicit DDRep  & \multicolumn{1}{r}{\textbf{66.9}} & \multicolumn{1}{r}{\textbf{66.8}} & \multicolumn{1}{r}{\textbf{61.6}} \\
Target-only &  \multicolumn{1}{r}{88.1}                          & \multicolumn{1}{r}{99.8}                            & \multicolumn{1}{r}{87.9} \\
\hline

\end{tabular}
  }
  \label{fmnist}%

\end{table}

\begin{table}[h]
\hfill
 \centering
  \caption{Z-test score value comparing MaxDIRep to other models on the constructed Fashion-MNIST dataset. z$>$2.3 means that the probability of MaxDIRep being no better is $\leq$0.01.}
  \vspace*{0.1cm} 

 \begin{tabular}{llll}
\hline
Model &
  \begin{tabular}[c]{@{}r@{}}No\\ cheating\end{tabular} &
  \begin{tabular}[c]{@{}r@{}}Shift\\ cheating\end{tabular} &
  \begin{tabular}[c]{@{}r@{}}Random\\ cheating\end{tabular} \\ \hline
Singla method~\citep{singla2020preparing} & \multicolumn{1}{r}{1.55}    & \multicolumn{1}{r}{3.28}  & \multicolumn{1}{r}{3.68}   \\
DANN~\citep{ganin2016domain}   &    \multicolumn{1}{r}{2.26}                      &    \multicolumn{1}{r}{4.17}                      &      \multicolumn{1}{r}{4.33}                        \\
DSN~\citep{bousmalis2016domain}       & \multicolumn{1}{r}{0.16} & \multicolumn{1}{r}{2.60} & \multicolumn{1}{r}{3.18}  \\
\hline

\end{tabular}
  
  \label{appz test FM}%

\end{table}

\subsubsection{The effect of single-bit DDRep}

\begin{figure}[t]
    \centering
    \includegraphics[width=0.85\linewidth]{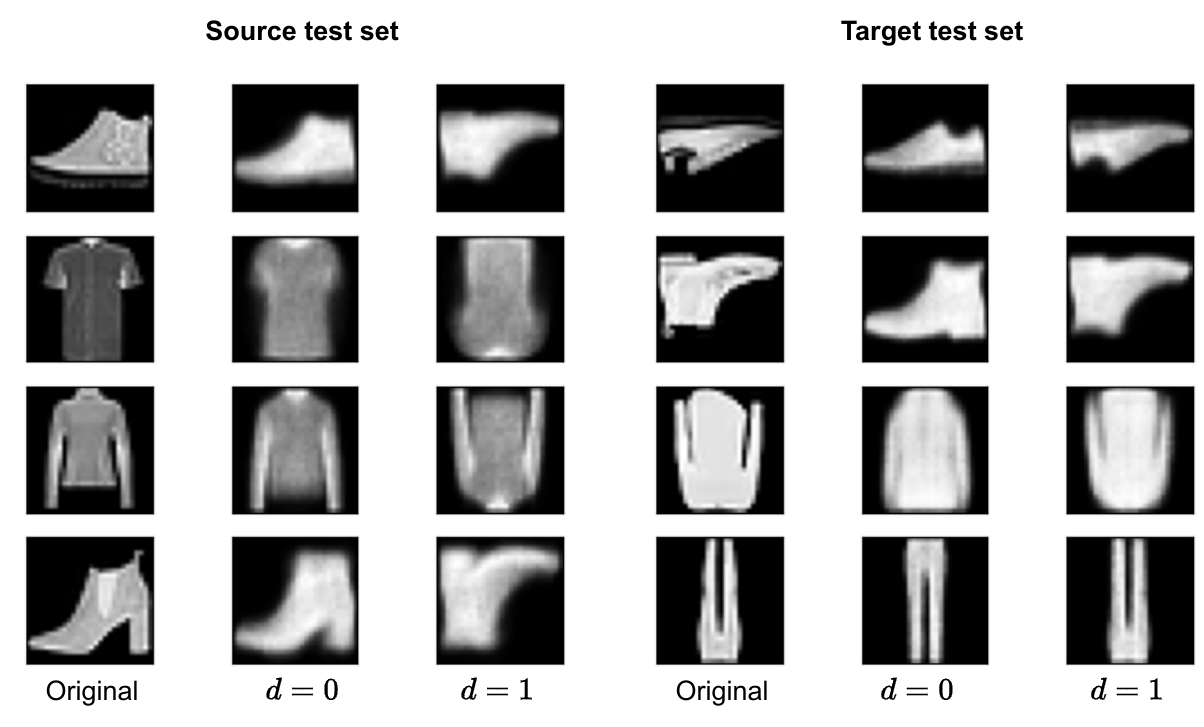}
    \caption{Effects of flipping the domain bit ($d \rightarrow 1-d$) on Fashion‑MNIST. Columns 1 and 4: original images; columns 2 and 6: reconstructions; columns 3 and 5:  reconstructions with flipped domain bit.}
    \label{recon_fm}
\end{figure}

A particularly useful feature of the explicit DDRep algorithm is that it allows us to directly examine the effect of the DDRep by flipping the domain bit ($d \rightarrow 1-d$). This is illustrated in Figure~\ref{recon_fm} for rotated Fashion-MNIST classification. The original source and target domain images are shown in columns 1 and 4, respectively. Reconstructed images with the domain bit $d$ set to reflect their corresponding domains (i.e., $d = 0$ for the source domain and $d = 1$ for the target domain) are shown in columns 2 and 6. Notably, flipping the domain bit ($d \rightarrow 1-d$), while keeping the same DIRep, yields images (columns 3 and 5) resembling those from the opposite domain. This demonstrates the effectiveness of employing the domain bit as the DDRep within the explicit DDRep model. Note that the domain bit is the simplest DDRep, since MaxDIRep's KL loss aims to match the DDRep to a standard Gaussian distribution.

\begin{figure}[t]
    \centering
    \includegraphics[width=0.85\linewidth]{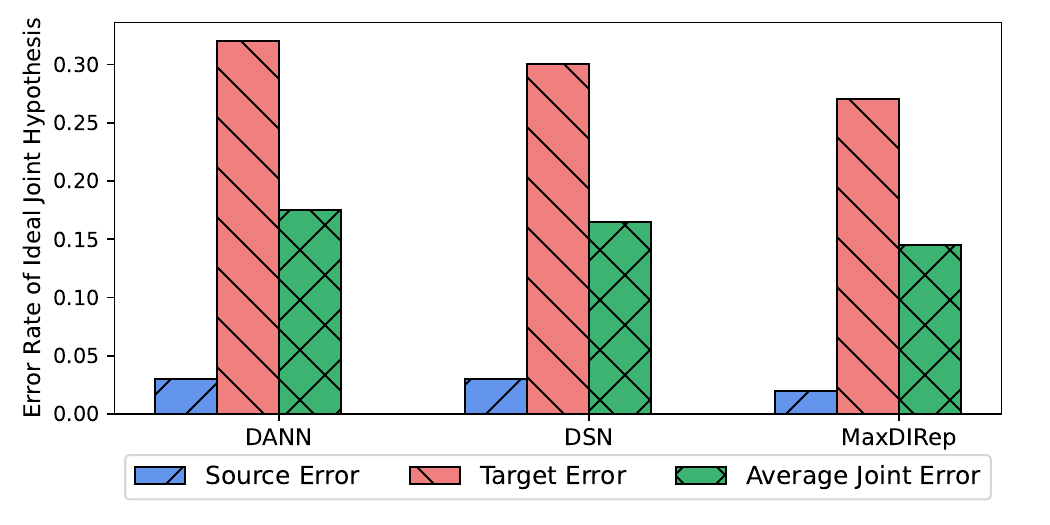}
    \caption{The error rate of the ideal joint hypothesis trained using representations learned by DANN, DSN, and MaxDIRep.}
    \label{error}
\end{figure}

\subsubsection{The error of an ideal joint hypothesis} We follow the same approach in the literature to find the ideal joint hypothesis~\citep{chen2019transferability, liu2019transferable} on this dataset. Specifically, we train a new MLP classifier using the DIReps learned by DANN,  DSN, and MaxDIRep, respectively. The MLP classifier is trained on both source and target training data with labels, while each DA model is fixed. The target labels are only used for evaluating the error of the ideal joint hypothesis and are not involved in training the DA models.   We then obtain the error rate of the trained MLP classifier on the source test set and target test set and calculate the average error rate.   The results in Figure~\ref{error} show that MaxDIRep achieves the lowest error rate for the ideal joint hypothesis across both domains, thereby establishing a lower error bound for the target domain as indicated by \textbf{Theorem 1}.

\subsection{Synthetic benchmark based on CIFAR-10}\label{synthetic_cifar10}

We are interested in a more natural DA scenario where the source and target images might be captured with different sensors and thus have different wavelengths and colors. To address this scenario, we create another cheating benchmark based on CIFAR-10 with different color planes. We introduce the cheating color plane, where the choice of the color planes in the source data has a spurious correlation with the labels, while such correlation is absent in the target domain.  Specifically, we create a source set with cheating color planes by encoding CIFAR-10 labels (0-9). For odd labels, only the blue channel is retained with probability ($p$), and either the blue or red channel is kept randomly for the rest. For even labels, only the red channel is retained with probability ($p$), and either the red or blue channel is kept randomly for the rest. The parameter ($p$) controls the spurious correlation strength between image color and label. In the target domain, only the green channel is retained for each CIFAR-10 image. We compare our approach with others using ($p$) values from $\{0,0.2,0.4,0.6,0.8,0.9,1.0\}$, where a larger ($p$) value indicates a higher spurious correlation, making DA more challenging.

\begin{figure}[t]
    \centering
    \includegraphics[width=\linewidth]{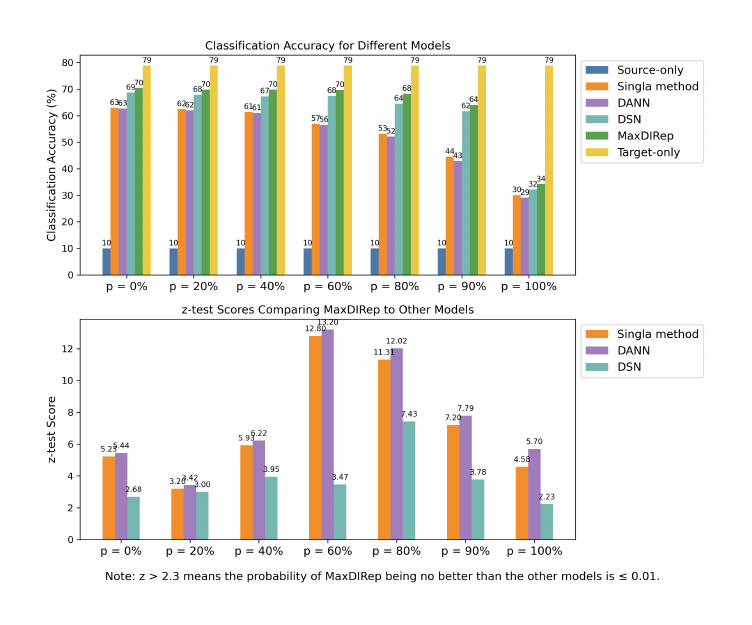}
    \caption{Classification accuracy (top) and z-test scores (bottom) for MaxDIRep and baseline models on CIFAR‑10 with varying probability $(p)$. The z-test compares MaxDIRep against DANN and DSN.}
    \label{cifar10}
\end{figure}

Figure~\ref{cifar10} presents the mean accuracy of MaxDIRep and the baseline algorithms on the target test set in a zero-shot setting.   We used the full MaxDIRep model due to its better performance. The
z-scores of the comparison of our method with other methods are shown in Figure~\ref{cifar10}. We observe similar performance degradation for the DANN, DSN and Singla method approaches on this benchmark, suggesting that the adaptation difficulties of previous methods and the better results achieved by our method are not limited to a particular dataset. Due to space limits, details of the experiments are given in Appendix~\ref{appcifar}. 

\begin{figure}[t]
\centering
\includegraphics[width=0.85\linewidth]{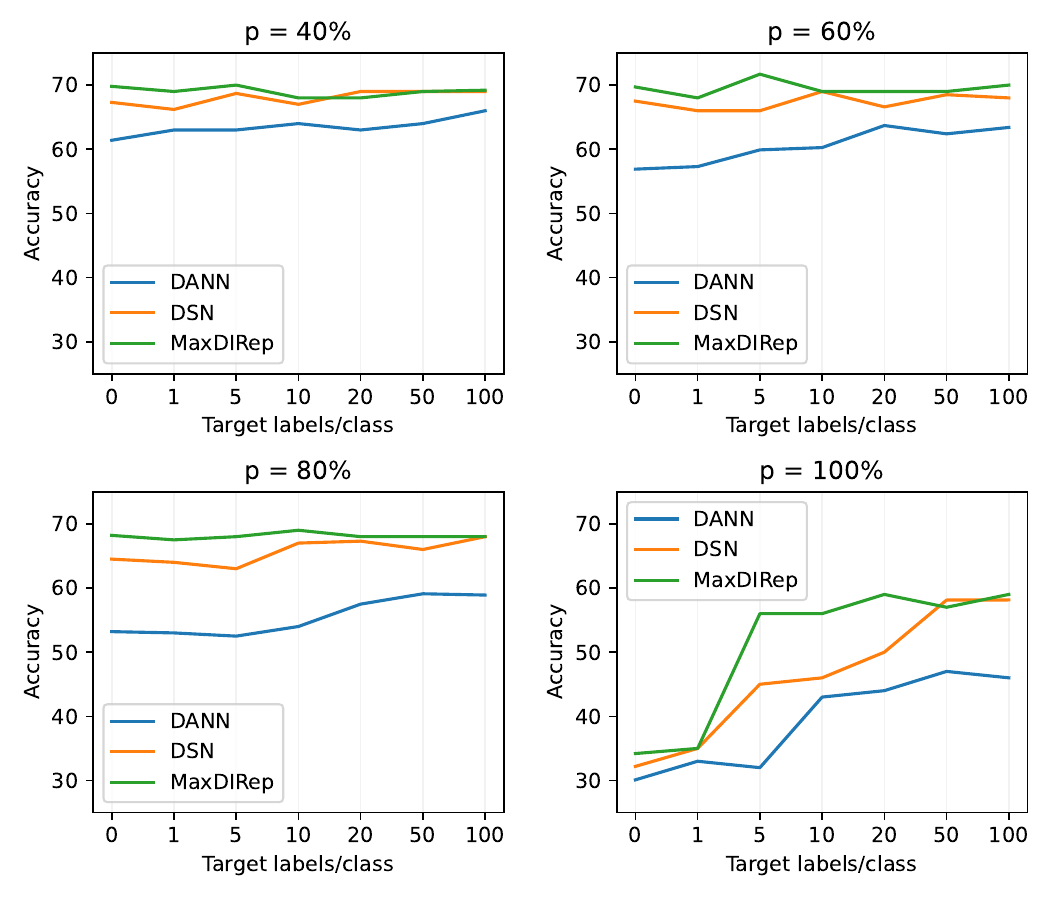}
\caption{Mean classification accuracy on CIFAR‑10 in the few‑shot setting for varying numbers of labeled target samples per class.}
\label{semi} 
\end{figure}

\subsubsection{Few-shot learning}To further evaluate MaxDIRep, we conduct experiments in a few-shot adaptation setting, where the model is provided with a majority of unlabeled target data and a small amount of labeled target data. We reveal $1, 5, 10, 20, 50,$ and $100$ labels per target class, which are incorporated into the classification loss through the label prediction pipeline. The same labeled samples are provided to DANN and DSN for a fair comparison. This setting allows us to examine how effectively each method leverages limited labeled target data under different spurious correlation levels. As shown in Figure~\ref{semi}, for $p$ values of $40\%$, $60\%$, and $80\%$, the classification accuracy improves moderately with additional labeled samples, while the performance order  MaxDIRep $>$ DSN $>$ DANN remains consistent. At $ p = 100\%$ , all methods achieve substantial gains; however, MaxDIRep achieves the highest improvement, surpassing DSN and DANN by $12\%$ and $25\%$, respectively, with only $50$ labeled target samples ($5$ labels per class). These results demonstrate that, even with limited labeled target data, MaxDIRep mitigates the influence of ``cheating information'' more effectively and has the best accuracy on the target set.

\subsection{The mutual ablation experiment between DSN and MaxDIRep}
\label{DIRep_max}

 In DSN, the orthogonality constraint is enforced by a difference loss ($\mathcal{L}_\mathit{diff}$), while minimizing the information content of DDRep in MaxDIRep is enforced by a KL loss ($\mathcal{L}_{kl}$) for the DDRep. To demonstrate the difference between DSN and MaxDIRep, we designed mutual ablation experiments to answer the following questions:
 \begin{itemize}
 \item If we add a negative difference loss ($-\mathcal{L}_\mathit{diff}$) to MaxDIRep, would the performance of MaxDIRep decrease? 
 \item On the other hand, if we add a KL loss for the DIRep ($\mathcal{L}^{DI}_\mathit{kl}$) in DSN, which acts as the opposite of the KL loss for the DDRep as in MaxDIRep, how would that affect the performance of DSN?    
\end{itemize}

In the two sets of ablation experiments (shaded blue and yellow, respectively, in Table~\ref{table:anti}), we perturb the systems by adding the KL loss for DIRep ($\lambda_{p}\mathcal{L}^{DI}_\mathit{kl}$) and the inverse difference loss ($-\lambda_p \mathcal{L}_{\mathit{diff}}$) to DSN and MaxDIRep, respectively. Here, $\lambda_{p}$ represents the strength of the perturbation. We use one large and one small value of $\lambda_p=0.001,\; 0.1$ (rows $2\&4$ for DSN, and rows $7\&9$ for MaxDIRep in Table~\ref{table:anti}) to explore the dependence on the perturbation strength. We then turn off these perturbations and continue the training until convergence to investigate if the systems can recover their original DA performance (rows $3\&5$ for DSN, and rows $8\&10$ for MaxDIRep in Table~\ref{table:anti}). For reference, we also list the performance when using the source data alone, DSN, and MaxDIRep in rows 1, 6, and 11, respectively, in Table~\ref{table:anti}.   

We provide values of the loss functions in the mutual ablation experiments. Table~\ref{appdsn} shows the effect of adding the KL loss for DIRep ($\lambda_{p}\mathcal{L}^{DI}_\mathit{kl}$) to DSN on DSN's loss functions. Table~\ref{appmax} shows the effect of adding the inverse difference loss ($-\lambda_p \mathcal{L}_{\mathit{diff}}$) to MaxDIRep on MaxDIRep's loss functions. 

\begin{table*}[h]
\centering
\caption{ Results of the ablation experiments conducted on the synthetic benchmark based on Fashion-MNIST (best viewed in color). Rows 2–5 show DSN perturbed with KL loss on DIRep, and rows 7–10 show MaxDIRep perturbed with negative difference loss. See text for details. }  
\resizebox{0.7\linewidth}{!}{
\begin{tabular}{lrrr}
\hline
Methods                                                 & \multicolumn{1}{l}{No cheating} & \multicolumn{1}{l}{Shift cheating} & \multicolumn{1}{l}{Random cheating} \\ \hline
1. Source only                                                                  & 20.0                            & 11.7                               & 13.8                                \\
\rowcolor[HTML]{D9DBFC} 
2. DSN + $\lambda_p\mathcal{L}^{DI}_{\mathit{kl}}$ ($\lambda_p = 0.001$)              & 61.2                            & 59.5                              & 53.8                                \\
\rowcolor[HTML]{D9DBFC} 
3. DSN* from 2                                                                 & 62.7                            & 60.3                               & 55.9                                \\
\rowcolor[HTML]{CBCEFB} 
4. DSN + $\lambda_p\mathcal{L}^{DI}_{\mathit{kl}}$ ($\lambda_p = 0.1$)                & 18.3                            & 12.7                               & 12.1                                \\
\rowcolor[HTML]{CBCEFB} 
5. DSN*  from 4                                                                 & 32.6                            & 29.7                               & 14.0                                \\

\cellcolor[HTML]{FFFFFF}6. DSN                                                  & 66.8                            & 63.6                               & 57.1                                \\
\rowcolor[HTML]{FFFFC7} 
7. MaxDIRep $-\lambda_p \mathcal{L}_\mathit{{diff}}$ ($\lambda_p = 0.001$) & \textbf{66.8}                            & \textbf{66.8}                      & 60.1                                \\
\rowcolor[HTML]{FFFFC7} 
8. MaxDIRep* from 7                                                             & \textbf{66.9}                   & \textbf{66.8}                      & 60.2                                \\
\rowcolor[HTML]{FFFE65} 
9. MaxDIRep $-\lambda_p \mathcal{L}_\mathit{{diff}}$ ($\lambda_p = 0.1$)   & 63.6                            & 63.6                               & 60.1                                \\
\rowcolor[HTML]{FFFE65} 
10. MaxDIRep* from 9                                                            & 65.5                            & 65.5                               & 60.3                                \\
11. MaxDIRep                                                                    & \textbf{66.9}                   & \textbf{66.8}                      & \textbf{61.6}                       \\ \hline
\end{tabular}
}

\label{table:anti}
\end{table*}

\begin{table*}[h]
\caption{Effect of adding the KL loss for DIRep $\lambda_{p}\mathcal{L}^{DI}_\mathit{kl}$ to DSN on DSN's loss functions. The loss values reported here are the average data from both the source and the target. }
\label{appdsn}
\centering
\resizebox{0.8\linewidth}{!}{
\begin{tabular}{@{}lrrrrrrrrr@{}}
\toprule
                                                                                           & \multicolumn{3}{c}{No cheating}                                                                                                                                                                                                       & \multicolumn{3}{c}{Shift cheating}                                                                                                                                                                                                    & \multicolumn{3}{c}{Random cheating}                                                                                                                                                                                                   \\ \cmidrule(l){2-10} 
\multirow{-2}{*}{Methods}                                                                  & \multicolumn{1}{c}{\cellcolor[HTML]{FFFFFF}$\mathcal{L}^{DI}_{\mathit{kl}}$ } & \multicolumn{1}{c}{\cellcolor[HTML]{FFFFFF}$\mathit{\mathcal{L}_{recon}}$} & \multicolumn{1}{c}{\cellcolor[HTML]{FFFFFF}$\mathit{\mathcal{L}_{diff}}$} & \multicolumn{1}{c}{\cellcolor[HTML]{FFFFFF}$\mathcal{L}^{DI}_{\mathit{kl}}$ } & \multicolumn{1}{c}{\cellcolor[HTML]{FFFFFF}$\mathit{\mathcal{L}_{recon}}$} & \multicolumn{1}{c}{\cellcolor[HTML]{FFFFFF}$\mathit{\mathcal{L}_{diff}}$} & \multicolumn{1}{c}{\cellcolor[HTML]{FFFFFF}$\mathcal{L}^{DI}_{\mathit{kl}}$ } & \multicolumn{1}{c}{\cellcolor[HTML]{FFFFFF}$\mathit{\mathcal{L}_{recon}}$} & \multicolumn{1}{c}{\cellcolor[HTML]{FFFFFF}$\mathit{\mathcal{L}_{diff}}$} \\ \midrule
\rowcolor[HTML]{FFFFFF} 
\cellcolor[HTML]{FFFFFF}2. DSN + $\lambda_p\mathcal{L}^{DI}_{\mathit{kl}}$ ($\lambda_p = 0.001$)  & 29.7                                                                         & 0.04                                                                       & 0                                                                         & 19.7                                                                         & 0.04                                                                       & 0                                                                         & 25.8                                                                         & 0.05                                                                       & 0                                                                         \\
\rowcolor[HTML]{FFFFFF} 
\cellcolor[HTML]{FFFFFF}3. DSN*  from 2                                                    & 41.5                                                                         & 0.04                                                                       & 0                                                                         & 48.6                                                                         & 0.04                                                                       & 0                                                                         & 30.6                                                                         & 0.05                                                                       & 0                                                                         \\
\rowcolor[HTML]{FFFFFF} 
\cellcolor[HTML]{FFFFFF}4. DSN + $\lambda_p\mathcal{L}^{DI}_{\mathit{kl}}$ ($\lambda_p = 0.1$)      & \cellcolor[HTML]{FFFFFF}1.725                                                & \cellcolor[HTML]{FFFFFF}0.05                                               & \cellcolor[HTML]{FFFFFF}0                                                 & 1.65                                                                         & 0.05                                                                       & 0                                                                         & 2.04                                                                         & 0.06                                                                       & 0                                                                         \\
\rowcolor[HTML]{FFFFFF} 
\cellcolor[HTML]{FFFFFF}5. DSN*  from 4                                                    & \cellcolor[HTML]{FFFFFF}16                                                   & \cellcolor[HTML]{FFFFFF}0.05                                               & \cellcolor[HTML]{FFFFFF}0                                                 & 14.3                                                                         & 0.04                                                                       & 0                                                                         & 11.9                                                                         & 0.06                                                                       & 0                                                                         \\
\rowcolor[HTML]{FFFFFF} 
\cellcolor[HTML]{FFFFFF}6. DSN                                                             & \cellcolor[HTML]{FFFFFF}N/A                                                  & \cellcolor[HTML]{FFFFFF}0.04                                               & \cellcolor[HTML]{FFFFFF}0                                                 & N/A                                                                          & 0.04                                                                       & 0                                                                         & N/A                                                                          & 0.05                                                                       & 0                                                                         \\ \bottomrule
\end{tabular}}
\end{table*}

\begin{table*}[h]
\caption{Effect of adding the inverse difference loss $-\lambda_p \mathcal{L}_{\mathit{diff}}$  to MaxDIRep on MaxDIRep's loss functions. The loss values reported here are the average data from both the source and the target. }
\label{appmax}
\centering
\resizebox{0.8\linewidth}{!}{
\begin{tabular}{@{}lrrrrrrrrr@{}}
\toprule
                                                                                                        & \multicolumn{3}{c}{No cheating}                                                                                                                                                                                                            & \multicolumn{3}{c}{Shift cheating}                                                                                                                                                                                                         & \multicolumn{3}{c}{Random cheating}                                                                                                                                                                                                        \\ \cmidrule(l){2-10} 
\multirow{-2}{*}{Methods}                                                                               & \multicolumn{1}{c}{\cellcolor[HTML]{FFFFFF}$\mathcal{L}_{\mathit{kl}}$} & \multicolumn{1}{c}{\cellcolor[HTML]{FFFFFF}$\mathit{\mathcal{L}_{recon}}$} & \multicolumn{1}{c}{\cellcolor[HTML]{FFFFFF}$\mathit{\mathcal{L}_{diff}}$} & \multicolumn{1}{c}{\cellcolor[HTML]{FFFFFF}$\mathcal{L}_{\mathit{kl}}$} & \multicolumn{1}{c}{\cellcolor[HTML]{FFFFFF}$\mathit{\mathcal{L}_{recon}}$} & \multicolumn{1}{c}{\cellcolor[HTML]{FFFFFF}$\mathit{\mathcal{L}_{diff}}$} & \multicolumn{1}{c}{\cellcolor[HTML]{FFFFFF}$\mathcal{L}_{\mathit{kl}}$} & \multicolumn{1}{c}{\cellcolor[HTML]{FFFFFF}$\mathit{\mathcal{L}_{recon}}$} & \multicolumn{1}{c}{\cellcolor[HTML]{FFFFFF}$\mathit{\mathcal{L}_{diff}}$} \\ \midrule
\rowcolor[HTML]{FFFFFF} 
\cellcolor[HTML]{FFFFFF}7. MaxDIRep $-\lambda_p \mathcal{L}_\mathit{{diff}}$ ($\lambda_p = 0.001$) & 0                                                                             & 0.07                                                                       & 0                                                                             & 0                                                                             & 0.07                                                                       & 0                                                                             & 0                                                                             & 0.07                                                                       & 0                                                                             \\
\rowcolor[HTML]{FFFFFF} 
\cellcolor[HTML]{FFFFFF}8. MaxDIRep* from 7                                                             & 0                                                                             & 0.07                                                                       & 0                                                                             & 0                                                                             & 0.07                                                                       & 0                                                                             & 0                                                                             & 0.07                                                                       & 0                                                                             \\
\rowcolor[HTML]{FFFFFF} 
\cellcolor[HTML]{FFFFFF}9. MaxDIRep $-\lambda_p \mathcal{L}_\mathit{{diff}}$ ($\lambda_p = 0.1$) & \cellcolor[HTML]{FFFFFF}0                                                     & \cellcolor[HTML]{FFFFFF}0.07                                               & \cellcolor[HTML]{FFFFFF}0                                                     & 0                                                                             & 0.07                                                                       & 0                                                                             & 0                                                                             & 0.07                                                                       & 0                                                                             \\
\rowcolor[HTML]{FFFFFF} 
\cellcolor[HTML]{FFFFFF}10. MaxDIRep* from 9                                                            & \cellcolor[HTML]{FFFFFF}0                                                     & \cellcolor[HTML]{FFFFFF}0.07                                               & \cellcolor[HTML]{FFFFFF}0                                                     & 0                                                                             & 0.07                                                                       & 0                                                                             & 0                                                                             & 0.07                                                                       & 0                                                                             \\
\rowcolor[HTML]{FFFFFF} 
\cellcolor[HTML]{FFFFFF}11. MaxDIRep                                                                    & \cellcolor[HTML]{FFFFFF}0                                                     & \cellcolor[HTML]{FFFFFF}0.07                                               & \cellcolor[HTML]{FFFFFF}N/A                                                   & 0                                                                             & 0.07                                                                       & N/A                                                                           & 0                                                                             & 0.07                                                                       & N/A                                                                           \\ \bottomrule
\end{tabular}}
\end{table*}

The finding in row 2 of Table~\ref{table:anti} indicates that when we minimize the information content in DIRep during DSN training, DDRep and DIRep maintain orthogonality as evidenced by $\mathcal{L}_\mathit{diff} = 0$ in the experiment (see Table~\ref{appdsn}). However, even this weak perturbation results in a worse DA performance than the original DSN. The results also show that even after this perturbation is removed (row 3), the optimal DA is not regained. This is consistent with the geometric analogy (Figure~\ref{DA_Geom}), which shows that many solutions satisfy the orthogonal constraint, but not all are equally good in DA. Here, DSN finds a sub-optimal solution from the initiation of weights reached by a weak ``ablation'' perturbation. Additionally, if we apply a stronger perturbation (row 4 in Table~\ref{table:anti}), the DSN algorithm becomes equivalent to a source-only DA scheme. Notably, the values for reconstruction loss and difference loss do not increase, and the classification loss on the source data is minimal (see the reported loss values in Table~\ref{appdsn}). This implies that DIRep predominantly carries the label information for the source and random information for the target, while DDRep retains the information necessary for reconstruction. Another important observation is that the KL losses on DIRep in the ablation experiments for DSN (rows $2\&3$) with the smaller perturbation strength ($\lambda_p=0.001$) are significantly larger than those with the stronger perturbation ($\lambda_p=0.1$, rows $4\&5$) (the loss values are reported in Table~\ref{appdsn}). This confirms that a better DA is achieved with a higher information content in DIRep. 

On the contrary, the performance of MaxDIRep is largely unaffected by the perturbation regardless of its strength (rows 7-10 in Table~\ref{table:anti}). This is because minimizing the information content of DDRep in MaxDIRep imposes a much stronger constraint, which contains the weaker orthogonal constraint imposed by $\mathcal{L}_\mathit{diff}$. This is additionally supported by the observation that $\mathcal{L}_\mathit{diff} = 0$ in the ablation experiments for MaxDIRep (see Table~\ref{appmax}).

\subsection{Standard DA image benchmarks}\label{Standard}

\subsubsection{Office-31 dataset}
\begin{figure}[t]
\center
\includegraphics[width=0.8\linewidth]{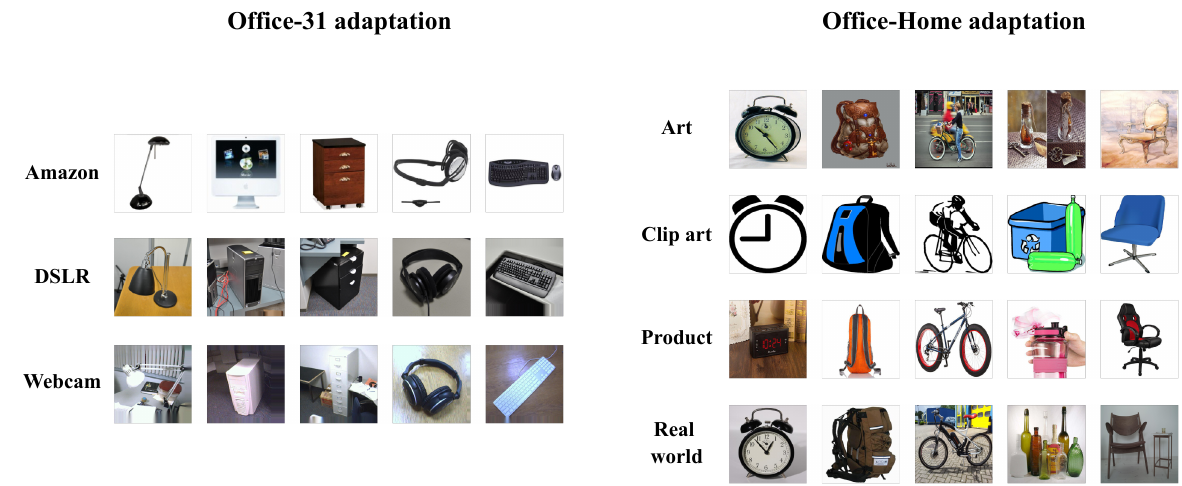}
\caption{Example images from different domains  in Office-31 and Office-Home.}
\label{officedemo} 
\end{figure}

The most commonly used dataset for DA in object classification is Office-31~\citep{saenko2010adapting}. The Office dataset has $4,110$ images from $31$ classes in three domains: Amazon ($2,817$ images), Webcam ($795$ images) and DSLR ($498$ images). Example images from all three datasets are provided in Figure~\ref{officedemo}. The three most challenging domain shifts reported in previous works are DSLR to Amazon ($D \rightarrow A$),  Webcam to Amazon ($W \rightarrow A$) and Amazon to DSLR ($A\rightarrow D$). $D \rightarrow A$ and $W \rightarrow A$  are the cases with the fewest labels in the source domain.

Following previous work~\citep{tzeng2017adversarial, chen2020adversarial}, we use a pretrained ResNet‑50 on ImageNet~\citep{deng2009imagenet} as the base model. Table~\ref{office} reports results for four zero‑shot adaptation tasks, where the full MaxDIRep model is used due to its superior performance. MaxDIRep is competitive on this adaptation task, matching the performance of CDAN~\citep{long2018conditional} in $A \rightarrow D$ and $W \rightarrow D$, and
outperforming all the approaches in all other tasks. However, it is worth noting that CDAN~\citep{long2018conditional} utilizes a conditional discriminator conditioned on the cross-covariance of domain-specific feature representations and classifier predictions, which has the potential to improve our results further. We leave exploring this possibility for future work.  Our approach shows the most significant performance improvements in scenarios such as $D \rightarrow A$ and $W \rightarrow A$, in which background information is present within the $D$ and $W$ domains while being absent in the $A$ domain.

\begin{table*}[t]
\caption{Averaged accuracy (\%) of different  DA approaches on the Office-Home dataset.} 
\label{officehome}
\centering
\resizebox{\linewidth}{!}{
\begin{tabular}{@{}lrrrrrrrrrrrrr@{}}
\toprule
Methods                                                & \multicolumn{1}{c}{Ar-Cl} & \multicolumn{1}{c}{Ar-Pr} & \multicolumn{1}{c}{Ar-Rw} & \multicolumn{1}{c}{Cl-Ar} & \multicolumn{1}{c}{Cl-Pr} & \multicolumn{1}{c}{Cl-Rw} & \multicolumn{1}{c}{Pr-Ar} & \multicolumn{1}{c}{Pr-Cl} & \multicolumn{1}{c}{Pr-Rw} & \multicolumn{1}{c}{Rw-Ar} & \multicolumn{1}{c}{Rw-Cl} & \multicolumn{1}{c}{Rw-Pr} & \multicolumn{1}{c}{Avg} \\ \midrule
Source-only                                            & 34.9                      & 50.0                      & 58.0                      & 37.4                      & 41.9                      & 46.2                      & 38.5                      & 31.2                      & 60.4                      & 53.9                      & 41.2                      & 59.9                      & 46.1                    \\
DANN~\citep{ganin2016domain}     & 45.6                      & 59.3                      & 70.1                      & 47.0                      & 58.5                      & 60.9                      & 46.1                      & 43.7                      & 68.5                      & 63.2                      & 51.8                      & 76.8                      & 57.6                    \\
CDAN~\citep{long2018conditional} & 49.0                      & 69.3                      & 74.5                      & 54.4                      & 66.0                      & 68.4                      & 55.6                      & 48.3                      & 75.9                      & 68.4                      & 55.4                      & 80.5                      & 63.8                    \\
MCD~\citep{saito2018maximum}      & 45.6                      & 60.9                      & 69.2                      & 50.8                      & 60.7                      & 60.5                      & 46.2                      & 44.0                      & 74.7                      & 62.6                      & 53.8                      & 77.5                      & 58.6                    \\
GPDA~\citep{kim2019unsupervised}  & 47.1                      & 62.0                      & 70.4                      & 53.6                      & 62.3                      & 60.9                      & 49.7                      & 47.2                      & 72.3                      & 63.7                      & 54.0                      & 78.6                      & 60.2                    \\
MaxDIRep                                               & \textbf{53.5}             & \textbf{71.1}             & \textbf{78.9}             & \textbf{54.9}             & \textbf{66.0}             & \textbf{68.8}             & \textbf{59.5}             & \textbf{48.7}             & \textbf{78.6}             & \textbf{69.5}             & \textbf{56.6}             & \textbf{80.8}             & \textbf{65.6}           \\ \bottomrule
\end{tabular}
}
\end{table*}

\begin{table}[h]
\centering
\caption{Mean classification accuracy (\%) of different baseline approaches on the Office-31 dataset. The results are cited from each study when available. The results of MCD~\citep{saito2018maximum} is cited from~\citep{ma2021adversarial}. We present our DSN replication results on the Office-31 dataset. Office-31 was not evaluated by the DSN authors. }
\label{office}
\resizebox{\linewidth}{!}{
\begin{tabular}{lrrrr}
\hline
Model                                                          & \multicolumn{1}{l}{$D \rightarrow A$} & \multicolumn{1}{l}{$W \rightarrow A$} & \multicolumn{1}{l}{$W \rightarrow D$} & \multicolumn{1}{l}{$A \rightarrow D$} \\ \hline
Source-only                                                    & 62.5                                  & 60.7                                  & 98.6                                  & 68.9                                  \\
DANN~\citep{ganin2016domain}             & 68.2                                  & 67.4                                  & 99.2                                  & 79.7                                  \\
ADDA~\citep{tzeng2017adversarial}        & 69.5                                  & 68.9                                  & 99.6                                  & 77.8                                  \\
CDAN~\citep{long2018conditional}         & 70.1                                  & 68.0                                  & \textbf{100.0}                        & \textbf{89.8}                         \\
GTA~\citep{sankaranarayanan2018generate} & 72.8                                  & 71.4                                  & 99.9                                  & 87.7                                  \\
SimNet~\citep{pinheiro2018unsupervised}  & 73.4                                  & 71.8                                  & 99.7                                  & 85.3                                  \\
MCD~\citep{saito2018maximum}                                                       & 71.0                                  & 67.2                                  & 98.4                                  & 84.1                                  \\
GPDA~\citep{kim2019unsupervised}                                                             & 72.3                                  & 68.8                                  & \textbf{100}                          & 85.5                                  \\
AFN~\citep{xu2019larger}                 & 69.8                                  & 69.7                                  & 99.8                                  & 87.7                                  \\
Chadha et al.~\citep{chadha2019improved} & 62.2                                  & - $\ $                                & -                                     & 80.9                                  \\
IFDAN-1~\citep{deng2021informative}     & 69.2                                  & 69.4                                  & 99.8                                  & 80.1                                  \\
DSN~\citep{bousmalis2016domain}          & 67.2                                  & 67.5                                  & 98.0                                  & 82.0                                  \\
MaxDIRep                                                       & \textbf{73.8}                         & \textbf{72.5}                         & \textbf{100.0}                        & 89.0                      \\ \hline
\end{tabular}
}
\end{table}

\subsubsection{Office-Home dataset}

Office-Home - a more difficult dataset than Office-31, consists of $15,500$ images in $65$ object classes, forming four extremely dissimilar domains: Artistic images (Ar), Clip Art (Cl), Product images (Pr), and Real-World images (Rw).  Example images from all four datasets are provided in Figure~\ref{officedemo}.   We use the same ResNet-50 network with the same training protocols and the hyperparameters from CDAN~\citep{long2015learning}. 

Strong results are also achieved on the Office-Home dataset as reported in  Table~\ref{officehome} for the MaxDIRep. In the evaluation of $12$ transfer tasks, MaxDIRep consistently outperforms DANN~\citep{ganin2016domain}, CDAN~\citep{long2018conditional}, MCD~\citep{saito2018maximum},  and GPDA~\citep{kim2019unsupervised}. 
The classification accuracy of the Office-Home dataset is lower compared to the Office-31 dataset. The four domains in Office-Home have more categories and greater visual dissimilarity, making adaptation more difficult.


\subsection{Application in network intrusion detection (NID)}\label{nid}

Beyond image classification, we evaluate MaxDIRep on network intrusion detection (NID).  NID datasets consist of network features extracted from both malicious and benign network traffic flows. An NID detector is trained on these datasets to predict whether an incoming network flow is benign or originates from a network attack. Because labeling network traffic is labor-intensive, DA provides a valuable approach.

Singla et al. \cite{singla2020preparing} demonstrated the use of DA for NID, transferring knowledge from a labeled source dataset (e.g., from a Wi-Fi network) to a target dataset with limited labels (e.g., from an IoT network). This enables leveraging existing labeled data to train effective NID models for new network environments.

\begin{figure}[h]
\centering
\includegraphics[width=0.85\linewidth]{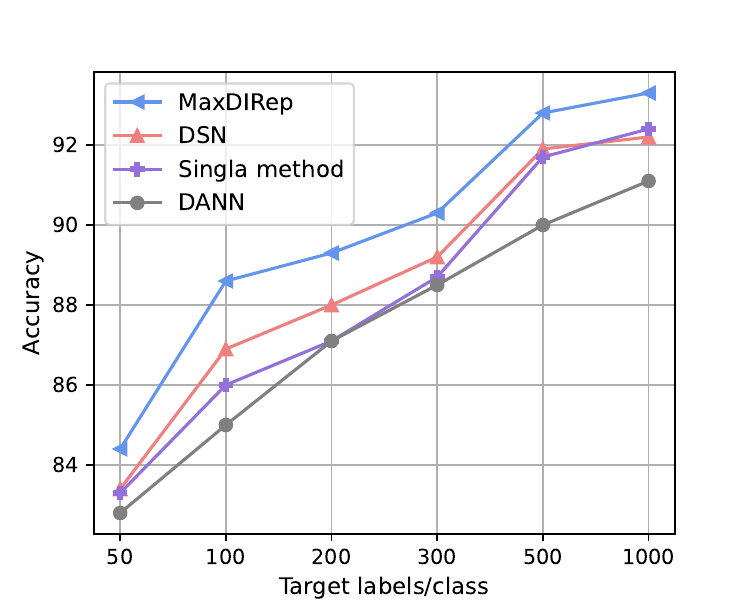}
\caption{Mean classification accuracy on UNSW-NB15 test-set in the few-shot setting.}
\label{nids} 
\end{figure}

Following Singla et al.~\cite{singla2020preparing}, we employ NSL-KDD \cite{nsltkdd} as the source dataset and UNSW-NB15 \cite{moustafa2015unsw} as the target dataset. NSL-KDD has $125,973$ samples, and UNSW-NB15 has $175,341$ samples. We evaluate MaxDIRep, DSN, and DANN using the same few-shot setting as Singla et al.~\cite{singla2020preparing}, training with all labeled source samples and varying the number of labeled target samples $\{50, 100, 200, 300, 500, 1000\}$ per class (benign and attack). As shown in Figure~\ref{nids}, all methods improve with increasing target labels, maintaining the following performance order: MaxDIRep $>$ DSN  $>$  Singla method $>$  DANN.

\section{Conclusion}\label{conclusion}

MaxDIRep achieves superior performance by ensuring that the DIRep retains target‑label‑relevant information. Unlike DSN’s weak orthogonality constraint or a discriminator alone, our KL loss on the DDRep prevents useful features from being discarded. Our claims are supported by our ablation experiments on a synthetic dataset and a geometrical analogy. We further validated MaxDIRep on an additional synthetic benchmark containing domain‑specific cues, where it again outperforms competing approaches. Across all standard DA benchmarks, MaxDIRep consistently surpasses recent DA methods. Finally, when adapted for network intrusion detection using source and target datasets from different networks with significant data drift, MaxDIRep again achieves superior results over prior approaches.

Future work could explore integrating pseudo-labeling, a powerful technique that uses pseudo-labels to provide noisy but sufficiently accurate labels for target data, enabling progressive model updates~\citep{chen2020adversarial,zou2018unsupervised}. While not addressed in this work, we anticipate that integrating pseudo-labeling with MaxDIRep would further enhance adaptation performance.\\

\noindent
{\bf Acknowledgments.} The work reported in this paper has been supported by the National Science Foundation (NSF) under Grants 2229876 and 2112471.

\bibliography{main}
\bibliographystyle{plain}

\appendices


\section{Experiment Details on Fashion-MNIST}\label{appfm}

\subsubsection{Network architecture}\label{appfm_tp}

All methods were trained using the Adam optimizer with a learning rate of $2e-4$ for $10,000$ iterations and batch sizes of $128$ samples per domain ($256$ total). MaxDIRep's label prediction pipeline (generator and classifier) consists of eight fully connected layers (\textsc{fc1}-\textsc{fc7}, \textsc{fc\_out}). Layers \textsc{fc1}-\textsc{fc4} each have $100$ neurons; \textsc{fc5} (DIRep) has $100$ units; \textsc{fc6} and \textsc{fc7} each have $400$ units; and \textsc{fc\_out} is the output layer. The discriminator and decoder each have four hidden layers with $400$ units each, followed by the domain prediction and reconstruction layers, respectively. The encoder has two hidden layers with $400$ units each, followed by $100$-unit $z_\text{mean}$, $100$-unit $z_\text{variance}$, and a sampling layer. Other models used the same architecture as MaxDIRep, where applicable. For the Singla method and DANN, the decoder and associated losses were disabled. For DSN, the same network architecture was used, and $\mathcal{L}_g$ was used for the similarity loss, with the shared and private encoders producing output vectors of the same dimensionality~\citep{bousmalis2016domain}.

\subsubsection{Hyperparameters}\label{apphyper}
Following previous work~\citep{ganin2016domain}, $\mathcal{L}_g$ was initialized to 0 and linearly increased to 1 according to $\lambda = \frac{2}{1 + \exp(-t)} - 1$, where $t$ is the training iteration. Other hyperparameters were set to $\beta = 0.1$, $\gamma = 0.15$, and $\mu = 0.1$ (without validation tuning). We closely followed the setup of loss function weights used in the DSN~\citep{bousmalis2016domain} and DANN~\citep{ganin2016domain} papers. To improve DSN's performance, we set the coefficient of $\mathcal{L}_{\mathrm{recon}}$ to 0.15 and the coefficient of $\mathcal{L}_{\mathrm{diff}}$ to 0.05, parameter values determined by~\citep{bousmalis2016domain} using a target label validation set. For a fair comparison, we used the same schedule for the coefficient of $\mathcal{L}_g$ and set the coefficient of $\mathcal{L}_c$ to 0.1 in DSN.

\section{Experiment details on CIFAR-10}\label{appcifar}


\subsubsection{Network architecture} 

We implement all network components using deep residual networks (ResNets) with shortcut 
connections~\cite{he2016deep}, as they are easier to optimize and can benefit from increased depth. Our implementation follows the full MaxDIRep architecture.

\textbf{Label prediction pipeline:} The classifier adopts a ResNet-20 backbone, as in CIFAR-10. The generator begins with a $3 \times 3$ convolutional layer, followed by a stack of $6$ residual blocks with $3 \times 3$ convolutions on feature maps of size $32$, using $16$ filters. The classifier consists of $6 \times 2$ residual blocks with $3 \times 3$ convolutions on feature maps of sizes $\{16, 8\}$, with $\{32, 64\}$ filters, respectively. A global average pooling layer and a fully connected softmax layer are appended for label prediction.

\textbf{Discriminator:} The discriminator takes $32 \times 32 \times 16$ domain-invariant features as input. It starts with a $3 \times 3$ convolutional layer, followed by $6 \times 3$ residual blocks with $3 \times 3$ convolutions on feature maps of sizes $\{32, 16, 8\}$, using $\{16, 32, 64\}$ filters, respectively. The network concludes with global average pooling, a fully connected layer with two outputs, and a softmax layer.

\textbf{Encoder and decoder:} The encoder consists of two shared convolutional layers: a $3 \times 3$ layer with $3$ filters, followed by a $3 \times 3$ layer with $2$ filters. The resulting feature maps are passed to two parallel branches, each containing a $3 \times 3$ convolutional layer with $2$ filters, to estimate the mean and log-variance of the latent variable. A sampling layer then generates the DDRep by drawing samples from the latent distribution parameterized by these two outputs. The decoder reconstructs the input image using both DIRep and DDRep. The configuration of the decoder is the inverse of that of the generator. 

For fair comparison, we apply the same ResNet-based architecture to all other approaches whenever applicable.

\subsubsection{Hyperparameters} We use a weight decay of $0.0001$ and adopt the BN~\citep{ioffe2015batch} for all the experiments.  We use the same schedule in Appendix \ref{apphyper} for the coefficient of $\mathcal{L}_g$ in all the experiments.  For other hyperparameters, we used $\beta=1, \gamma = 1, \mu=1/2000$ in MaxDIrep and  set the coefficient of $\mathcal{L}_{\mathit{recon}}$ to $0.15$, the coefficient of $\mathcal{L}_{\mathit{diff}}$ to $0.05$, and the coefficient of $\mathcal{L}_{\mathit{c}}$ to $1$ in DSN.

\section {Proof for the Geometrical Analogy}\label{geom}

To understand the difference between DSN and MaxDIRep, we looked at a 3-D geometrical interpretation of representation decomposition as shown in Figure~\ref{DA_Geom}. Here, we show that all points on the blue circle satisfy the orthogonal condition, i.e., $\mathit{DI_D\perp DD_D^{S,T}}$.

The source and target data are represented by two vectors $S=\vv{OS}$, $T=\vv{OT}$ where $O$ is the origin, as shown in Figure~\ref{DA_Geom}. We assume the source and target vectors have equal amplitude $|\vv{OS}|=|\vv{OT}|$.  Let us define the plane that passes through the triangle $O-S-T$ as plane-$\mathcal{A}$ (the gray plane in Figure~\ref{DA_Geom}). The mid-point between $S$ and $T$ is denoted as $V$. Let us draw another plane (the blue plane-$\mathcal{B}$) that passes through the line $OV$ and is perpendicular to the plane-$\mathcal{A}$. The blue circle is on the blue plane-$\mathcal{B}$ with a diameter given by $OV$. Denote an arbitrary point on the blue circle as D with the angle $\angle \mathit{DVO}=\theta$. Let us define the plane that passes through the triangle $D-S-T$ as plane-$\mathcal{C}$ (not shown in Figure~\ref{DA_Geom}).

Since the blue plane-$\mathcal{B}$ is the middle plane separating S and T, we have $ST\perp OV$ and $ST\perp DV$  (note that $XY$ represents the line between the two points $X$ and $Y$). Therefore, the line $ST$ is perpendicular to the whole plane-$\mathcal{B}$: $ST\perp\mathcal{B}$, which means that $ST$ is perpendicular to any line on plane-$\mathcal{B}$. Since the line $DV$ is on the plane-$\mathcal{B}$, we have $\mathit{OD}\perp ST$. Since $OV$ is the diameter of the blue circle, we have $\mathit{OD}\perp DV$. Since $DV$ and $ST$ span the plane-$\mathcal{C}$, we have $\mathit{OD}$ is perpendicular to the whole plane-$\mathcal{C}$: $\mathit{OD}\perp\mathcal{C}$, which means that $\mathit{OD}$ is perpendicular (orthogonal) to any line on plane-$\mathcal{C}$ including $DS$ and $DT$. Therefore, we have proved: $\mathit{OD}\perp DS$, $\mathit{OD}\perp DT$.

Note that with the notation given here, we can express the DIRep and DDRep for MaxDIRep (V) and DSN (D) as $$DI_V=\vv{OV}, \;\; \mathit{DD^S_V}=\vv{\mathit{VS}},\;\; \mathit{DD^T_V}=\vv{\mathit{VT}}.$$
$$DI_D=\vv{\mathit{OD}}, \;\; \mathit{DD^S_D}=\vv{\mathit{DS}},\;\; \mathit{DD^T_D=\vv{DT}}.$$

Since we have proved that $\mathit{OD}\perp DS$, $\mathit{OD}\perp DT$ for any point $D$ on the blue circle, this means that any point on the blue circle satisfies the orthogonality constraint $\mathit{DI_D\perp \mathit{DD_D^{S,T}}}$. 

 In MaxDIRep, the size of DDRep's, i.e., 
 $\mathit{||S-DI||+||T-DI||=(||\vv{\mathit{VS}}||^2+||\vv{DV}||^2)^{1/2}+(||\vv{\mathit{VT}}||^2}$\\$+||\vv{DV}||^2)^{1/2}$
 is minimized leading to an unique solution $DI_V$ shown as the red dot (point $V$) in Figure~\ref{DA_Geom}, which satisfies the orthogonality constraint ($\mathit{DI_V\perp DD_V^{S,T}}$) as it is on the blue circle. More importantly, the MaxDIRep solution is unique as it maximizes the magnitude of DIRep ($\mathit{||DI_V||\ge ||DI_D||}$). This can be seen easily as follows. Given the angle $\angle \mathit{DVO}=\theta$, we have $||DI_D||=||DI_V||\sin\theta \le ||DI_V||$.

\begin{table}[h]
\caption{We report the KL divergence ($\mathcal{L}_{kl}$) from our experiments, calculated as the average over data from both the source and target domains.}
\label{appkl}
\centering
\resizebox{\linewidth}{!}{
\begin{tabular}{@{}lrlr@{}}
\toprule
Task                            & \multicolumn{1}{c}{\begin{tabular}[c]{@{}c@{}}KL \\ divergence ($\mathcal{L}_{kl}$)\end{tabular}} & Task                               & \multicolumn{1}{c}{\begin{tabular}[c]{@{}c@{}}KL \\ divergence ($\mathcal{L}_{kl}$)\end{tabular}} \\ \midrule
Fashion-MNIST (no cheating)     & \multicolumn{1}{r|}{9.53e-07}                                                                     & Office-31 (W $\rightarrow$ D)      & 0.03                                                                                              \\
Fashion-MNIST (shift cheating)  & \multicolumn{1}{r|}{1.25e-06}                                                                     & Office-31 (A $\rightarrow$ D)      & 0.05                                                                                              \\
Fashion-MNIST (random cheating) & \multicolumn{1}{r|}{1.13e-06}                                                                     & Office-Home (Ar $\rightarrow$ Cl)  & 0.13                                                                                              \\
Office-31 (D $\rightarrow$ A)   & \multicolumn{1}{r|}{0.07}                                                                         & Office-Home (Ar $\rightarrow$ Rw)  & 0.10                                                                                              \\
Office-31 (W $\rightarrow$ A)   & \multicolumn{1}{r|}{0.03}                                                                                            & Office-Home (Rw $\rightarrow$  Cl) & 1                                                                                                 \\ \bottomrule
\end{tabular}
}
\end{table}

\end{document}